# SHADOWED AHP FOR Multi-criteria SUPPLIER SELECTION

Mohamed Abdel Hameed El-Hawy [1]

[1] Business information systems, Modern Academy for Computer Science and Management Technology, Cairo, Egypt

## ABSTRACT

*Numerous techniques of multi-criteria decision-making (MCDM) have been proposed in a variety of business domains. One of the well-known methods is the Analytical Hierarchical Process (AHP). Various uncertain numbers are commonly used to represent preference values in AHP problems. In the case of multi-granularity linguistic information, several methods have been proposed to address this type of AHP problem. This paper introduces a novel method to solve this problem using shadowed fuzzy numbers (SFNs). These numbers are characterized by approximating different types of fuzzy numbers and preserving their uncertainty properties. The new Shadowed AHP method is proposed to handle preference values which are represented by multi-types of uncertain numbers. The new approach converts multi-granular preference values into unified model of shadowed fuzzy numbers and utilizes their properties. A new ranking approach is introduced to order the results of aggregation preferences. The new approach is applied to solve a supplier selection problem in which multi-granular information are used. The features of the new approach are significant for decision-making applications.*

## 1. INTRODUCTION

Multiple techniques are developed to solve multi-criteria decision-making problems. When criteria evaluation is subjective and non-numeric, different types of fuzzy sets have been proposed to address this problem. The Analytic Hierarchy Process AHP method is one of the most important decision-making techniques that formalizing problems using a hierarchical structure. The basis of this methodology is the development of a pair-wise comparison between criteria and, similarly, between alternatives. In the literature, many approaches have been proposed to solve this problem using different types of uncertain numbers. The type-1 fuzzy numbers are the earliest form used for representing vague information and have been ap-plied in pair-wise comparisons for AHP problems. The first work is presented by Van Laarhoven et al. [1] which uses triangular fuzzy numbers. Buckley [2] developed fuzzy priorities for comparison values whose membership functions are trapezoidal. Chang [3] proposed a new approach for solving fuzzy AHP and used triangular fuzzy numbers. He presented the extent analysis method for the synthetic extent values of the pairwise comparisons. In the case of the supplier selection problem, different approaches are proposed. For example, Kahraman et al. [4] and Kilincci et al. [5] used the extent analysis method for fuzzy AHP to select the best supplier firms. Yang et al. [6] proposed a new approach using interpretive structural modeling (ISM) to map out the relationships among the sub-criteria and fuzzy AHP to compute the relative weights for each criterion. Chen et al. [7] and Sun [8] use two integrated approaches fuzzy AHP and fuzzy TOPSIS to select the best supplier. Pitchipoo et al. [9] combined two methods: fuzzy AHP for processing pairwise comparison matrices and grey relational analysis (GRA) for ranking sup-pliers. Ayhan et al. [10] developed a new approach that processes in two stages. In the first stage, the relative weights of each criterion are obtained by fuzzy AHP. In the next stage, the output of the previous step is the input of the Mixed Integer

Linear Programming (MILP) model which determines the suppliers and the quantities to be provided. In the study by Kar [11], the hybrid technique is applied, which combines fuzzy AHP and neural networks to provide group decision support under consensus achievement. AHP methods are developed with higher types of fuzzy sets. The studies of Görener et al [12] and Celik et al. [13] are examples of interval type-2 AHP. Xu et al. [14] and Duleba et al., [15] use intuitionistic fuzzy numbers in AHP problems.

The AHP method becomes more complex when comparisons are made using linguistic terms with different granular meanings. In the literature, some approaches have been proposed to solve this problem. Alsawy et al. [16] developed the new granular model, which considers a unified granular number to represent multi-granular numbers. Chatterjee et al. [17] applied these numbers with a fusion approach and used them with the VIKOR model. In the study by Aydin et al. [18], the transform technique is developed to convert three types of numbers (crisp, interval, and triangular fuzzy numbers) to standardized trapezoidal fuzzy numbers (STFN). In this paper, we will use the shadowed fuzzy numbers concept to deal with the multi-granular numbers problem. The multi-granular numbers are used for pair wise matrices. Multi-granular preferences include exact numbers, interval numbers, type-1 fuzzy numbers with two different types (triangular and Gaussian), and intuitionistic fuzzy numbers. To handle different granular numbers, the shadowed fuzzy numbers are used to unify different uncertain numbers. The new shadowed AHP algorithm is applied to solve supplier selection problems. The rest of this paper is organized as follows: Section 2 introduces the basic definitions of fuzzy sets, fuzzy numbers, shadowed sets, and shadowed fuzzy numbers (SFNs). In section 3, we introduce the proposed shadowed AHP method. Section 4, we present a numerical example that solves the AHP multi-granular supplier selection problem. In Section 5, the main features of the pro-posed approach are discussed. Finally, conclusions are presented in Section 6.

## 2. DEFINITIONS AND PRELIMINARIES

### 2.1. Fuzzy sets

Fuzzy sets are used to model linguistic terms by introducing gradual memberships. The membership function of a fuzzy set B is defined as follows [19]:

$$B: X \to [0, 1] \quad (1)$$

A fuzzy number (FN) $\widetilde{B}$ is a fuzzy set defined on the real numbers scale $\mathbb{R}$ with the following conditions [20, 21, 31].
1. $\widetilde{B}$ is normal, i.e. at least one element $x_i$ such that $\mu(x_i) = 1$.
2. $\widetilde{B}$ is convex such that $\widetilde{B}(\lambda x + (1 - \lambda)y) \geq \min\left(\widetilde{B}(x), \widetilde{B}(y)\right) \forall x, y \in U$ and $\lambda \in [0,1]$ where U is a universe of discourse.
3. The support of $\widetilde{B}$ is bounded.

A fuzzy number is important to approximate uncertainty concept about numbers or intervals. The membership function of the real fuzzy number $\widetilde{B}$ is defined by

$$\mu_{\widetilde{B}}(x) = \begin{cases} l_{\widetilde{B}}(x) & \text{if } a \leq x \leq b \\ 1 & \text{if } b \leq x \leq c, \\ r_{\widetilde{B}}(x) & \text{if } c \leq x \leq d, \\ 0 & \text{otherwise} \end{cases} \quad (2)$$

where $l_{\widetilde{B}}$ and $r_{\widetilde{B}}$ are two continuous increasing and decreasing functions for left and right side of fuzzy number and a, b, c, d are real numbers.

## 2.2. Shadowed sets

Shadowed sets are information granules constructed from fuzzy sets. These sets capture the essence of fuzzy sets, reducing computational costs and increasing the ability for interpretation. Shadowed set S defined as [22, 23]

$$S : X \rightarrow \{0, 1, [0,1]\} \quad (3)$$

When constructing a shadowed set, the fuzzy set membership function is replaced with a form similar to three-valued logic. The first step is selecting the threshold $\alpha \in (0, 0.5)$ to induce a shadowed set. The outcome of this step is the construction of three regions, as illustrated in Figure 2. The first region is reformulated by decreasing all membership values less than the threshold $\alpha$ to 0. The second region is constructed by elevated membership values greater than $1-\alpha$ to 1. The third region is unknown membership values or shadow regions for membership values around 0.5 [23] as illustrated in Figure 1. Selecting $\alpha$ is related to achieving a balance of uncertainty between the shadow region and the uncertainty lost by elevating and reducing the membership values to build the core and exclusion areas, as illustrated in the following equation [22].

$$v(r_1) + v(r_2) = v(r_3) \quad (4)$$

where $v$ is uncertainty of regions $r_1, r_2, r_3$. This balance is achieved by minimizing the performance index for the threshold $\alpha$ as shown in the following equation:

$$V_\alpha = |v(r_1) + v(r_2) - v(r_3)| \quad (5)$$

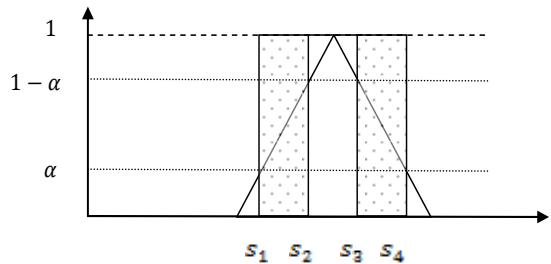

Figure 1. Shadowed sets induced from triangular fuzzy number

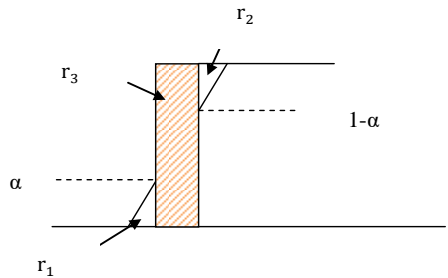

Figure 2. Regions that construct shadowed sets

## 2.3. Shadowed fuzzy numbers

Shadowed fuzzy number (SFN) is a special type of shadowed set. It is derived from fuzzy numbers [24]. The author developed an improved method to construct SFN that preserves the uncertainty characteristics of fuzzy numbers and can be induced from type-1 and higher types of fuzzy numbers, e.g., intuitionistic fuzzy sets (IFS) [25]. The author's approach can deduce SFN by building core interval and fuzziness intervals [26], as shown in Figure 3. The α-core for type-1 fuzzy numbers can be obtained by applying the following equation [26]

$$B_R(\alpha) - B_L(\alpha) + 1 = 2^{H_B} \tag{6}$$

where $H_B$ is the non-specificity value of fuzzy number B [26, 27]. $B_L(\alpha)$ and $B_R(\alpha)$ are left and right α-cut functions of B. The α-core interval is illustrated in Figure 3.

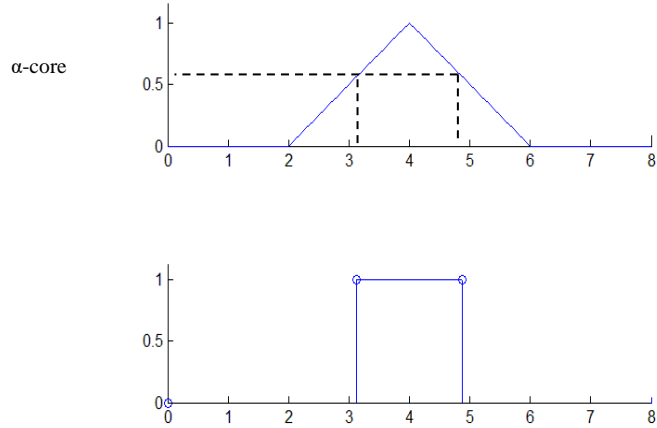

Figure 3. Core interval

The second step, the cardinality of the left and right fuzziness sets is used to generate the fuzziness intervals as

$$w_L = \sum_{x_L}^{\square} f_B(x_L) \tag{7}$$

$$w_R = \sum_{x_R}^{\square} f_B(x_R) \tag{8}$$

where $w_L, w_R$ are width of left and right of fuzziness intervals. $f_B(x_L), f_B(x_R)$ are fuzziness sets of fuzzy number that defined as the following[28].

$$f_B = (x, fuzz(x)), \tag{9}$$
$$fuzz(x) = 1 - |2\mu_B(x) - 1|. \tag{10}$$

Then shadowed fuzzy number (SFN) is induced by using core interval and fuzziness intervals as in Figure 4.
To extend this idea to intuitionistic fuzzy numbers (IFN), the author proposed the following steps [25]. Let A is an IFN defined on real numbers as [29]

$$A = \{< x, \mu_A(x), v_A(x) > | x \in \mathbb{R}\} \tag{11}$$

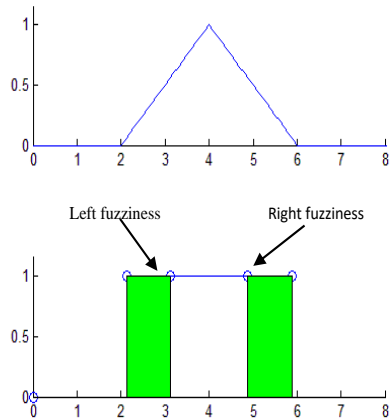

Figure 4. SFN for triangular fuzzy number

where $\mu_A(x) : X \to [0, 1]$ is membership function, $v_A(x) : X \to [0, 1]$ is non-membership function. The core interval of IFN is extracted using the following steps. The intuitionistic fuzzy number is transformed to interval fuzzy number as.

$$A(x) = [\mu(x), 1 - v(x)] \qquad (12)$$

where $\mu(x)$ is lower membership function and $1 - v(x)$ is upper membership function. Hartley non-specificity value of IFN is defined as

$$H_{avg}(A) = \frac{H(\mu(x)) + H(1 - v(x))}{2} \qquad (13)$$

where $H_{avg}$ the average of the non-specificity measure, $H(\mu(x))$ and $H(1 - v(x))$ are the Hartley non-specificity measure of the membership function and the complement of non-membership function. Then $H_{avg}$ is used to devise α-core for the lower membership function of intuitionistic fuzzy number (IFN)as the following equation [25].

$$A_r^L(\alpha) - A_l^L(\alpha) + 1 = 2^{H_{avg}} \qquad (14)$$

where $A_l^L(\alpha), A_r^L(\alpha)$ are the left and right α-cut equations of the lower membership function for IFN A. Then $\alpha_L$ is used in previous left and right α-cut equations to get the core interval $[c_l^L, c_r^L]$. Second, α-core is also deduced for the upper membership function of IFN $(\alpha_U)$ as shown in the following equation

$$A_r^U(\alpha) - A_l^U(\alpha) + 1 = 2^{H_{avg}} \qquad (15)$$

where $A_l^U(\alpha), A_r^U(\alpha)$ are the left and right α-cut of the upper membership function for IFN A. Also, by using $\alpha_U$ to induce the core interval of the higher membership function $[c_l^U, c_r^U]$. In the case of the symmetric membership function and the non-membership function, two core intervals are equal. In the case of non-symmetric, the average of two intervals is calculated as the following [25].

$$[c_l, c_r] = [\frac{(c_l^L + c_l^U)}{2}, \frac{(c_r^L + c_r^U)}{2}] \tag{16}$$

The shadow areas are obtained by using entropy intervals. Entropy is used to estimate the fuzziness of fuzzy sets. The intuitionistic fuzzy sets include both the fuzziness and the intuitionism of IFS [25, 30]. The cardinality of the left and right entropy set is used to get the wide of left and right entropy intervals as follows:

$$w_L = \sum_{x_L} E(A) \tag{17}$$

$$w_R = \sum_{x_R} E(A) \tag{18}$$

where $w_L$ and $w_R$ are the wide of left and right entropy intervals and $E(A)$ is entropy set for IFN A. It is defined as the following [25].

$$E(A) = (x, ent(x)) \tag{19}$$

$$ent(x) = \frac{1 - |\mu_A(x) - v_A(x)| + \pi_A(x)}{1 + \pi_A(x)} \tag{20}$$

where $ent(x)$ is entropy measure based on Xia Liang method [30] and $\mu_A$, $v_A$ and $\pi_A$ represent membership, non-membership, and hesitancy degree functions for every element belong to IFN. The core interval and entropy intervals are used to build the shadowed intuitionistic fuzzy numbers (SIFN), as shown in Figure 5.

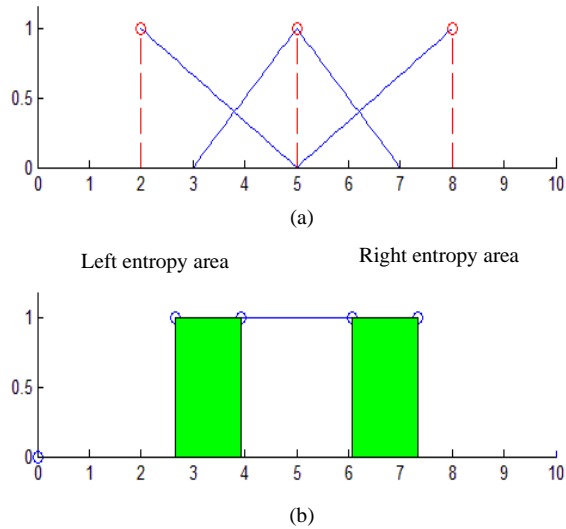

Figure 5: (a) IFN (b) SIFN

## 2.3.1. Shadowed fuzzy numbers operations

To use shadowed numbers in decision making applications, the basic arithmetic operations can be defined according to interval arithmetic operations [31] as the following:

Let $S(A) = (s_1^1, s_2^1, s_3^1, s_4^1)$, $S(B) = (s_1^2, s_2^2, s_3^2, s_4^2)$ are two shadowed fuzzy numbers (SFNs) that are induced from two fuzzy numbers A and B.

1) Addition

The addition operation of the two SFNs is defined as

$$S(A) + S(B) = (s_1^1 + s_1^2, s_2^1 + s_2^2, s_3^1 + s_3^2, s_4^1 + s_4^2) \qquad (21)$$

2) Subtraction

The subtraction operation of the two SFNs where $S(A) \geq S(B)$ is defined as the following

$$S(A) - S(B) = (s_1^1 - s_4^2, s_2^1 - s_3^2, s_3^1 - s_2^2, s_4^1 - s_1^2) \qquad (22)$$

3) Multiplication

The multiplication operation of the two SFNs where $S(A)$, $S(B)$ induced from two positive fuzzy numbers is defined as

$$S(A) \times S(B) = (s_1^1 \times s_1^2, s_2^1 \times s_2^2, s_3^1 \times s_3^2, s_4^1 \times s_4^2) \qquad (23)$$

4) Division

The division operation between the two SFNs $(S(A), S(B))$, where $s_1^2 \neq 0$, $s_2^2 \neq 0, s_3^2 \neq 0$ and $s_4^2 \neq 0$ and they induced from two positive fuzzy numbers is defined as the following

$$S(A)/S(B) = (s_1^1/s_4^2, s_2^1/s_3^2, s_3^1/s_2^2, s_4^1/s_1^2) \qquad (24)$$

## 3. THE PROPOSED SHADOWED AHP METHOD

In this section, we present a new shadowed AHP method. The proposed method aims to convert muti-granular input information for decision-making problems into shadowed fuzzy numbers and preserve the uncertainty characteristics of inputs. The following are the steps of the proposed approach:

### 3.1. Constructing comparisons matrices step:

The pairwise comparison matrices between criteria and also between alternatives are constructed by using the linguistic values of different types of uncertain numbers from 1 to 9 as in Table 1. These types include interval numbers, type-l fuzzy numbers (triangular TFNs and Gaussian GFNs) and intuitionistic fuzzy numbers (IFNs). Let $A = \{A_1, A_2, ..., A_i, ..., A_m\}$ is a set of alternatives, and $C = \{c_1, c_2, ..., c_j, ..., c_n\}$ is a set of criteria. We use the crisp numbers in Table 1 to calculate the consistency ratio for the multi-granular reciprocal matrices.

The scale in Table 1 from low to high preference represents equally important, moderately important, strongly important, very strongly important, and extremely important for pairwise comparison linguistic values.

**Table 1**: Preference scale with different types of uncertain numbers

| Crisp | interval | GFNs | TFNS | IFNs |
|---|---|---|---|---|
| 1 | [1, 2] | (1,0.5) | (1,1,2) | (0.5,1,2), (0.2,1,3) |
| 3 | [1, 5] | (3, 0.9) | (1, 3, 5) | (2,3,4), (1,3,5) |
| 5 | [3 , 7] | (5, 0.9) | (3, 5, 7) | (4,5,6), (3,5,7) |
| 7 | [5, 9] | (7, 0.9) | (5, 7, 9) | (6,7,8), (5,7,9) |
| 9 | [8, 10] | (9, 0.5) | (8, 9, 10) | (8,9,9.5), (7,9,10) |

2, 4, 6, and 8 are intermediate values

### 3.2. Unified uncertain numbers step:

Different types of fuzzy numbers are transformed to SFNs using operations defined by (6), (7), (8), (9), and (10) and (13), (14), (15), (16), (17), (18), (19), and (20). After the transformation process, a new reciprocal pairwise comparison matrix $R_{ij}^S$ is created as the following

$$R_{ij}^S = \begin{bmatrix} r_{11}^S & r_{12}^S & \cdots & r_{1n}^S \\ r_{21}^S & r_{22}^S & \cdots & r_{2n}^S \\ \vdots & \vdots & \vdots & \vdots \\ r_{n1}^S & r_{n2}^S & r_{nj}^S & r_{nn}^S \end{bmatrix} \quad (25)$$

where $r_{ij}^S = (S_1, S_2, S_3, S_4)$ is a shadowed fuzzy number (SFN) and $(i = 1, 2, \ldots, n, j = 1, 2, \ldots, n)$. In the case of alternatives matrix $(i = 1, 2, \ldots, m, j = 1, 2, \ldots, m)$.

### 3.3. The shadowed geometric mean step:
The shadowed geometric mean is computed as

$$g_i = (r_{i1}^S \times r_{i2}^S \times \ldots \times r_{in}^S)^{1/n} \quad (26)$$

The multiplication operation between SFNs is applied by using (23). Also, we use geometric mean with alternatives matrices.

### 3.4. The normalized weight step:
The normalized weight is obtained using the following equation

$$wg_i = \frac{g_i}{\sum_{i=1}^n g_i} \quad (27)$$

The addition and division operations between SFNs are used as (21) and (24). Also, the normalized priority of alternatives is obtained using (26) and (27).

### 3.5. The final preference values of alternatives step:

In this step, the final preference is calculated as

$$P_i = wg_i \times wa_i \tag{28}$$

where $wg_i$ is normalized weight of criteria and $wa_i$ is normalized priority of alternative.

### 3.6. The SFNs ranking step:

The result preference values of alternatives are SFNs that can be ranked using the following approach

Let $P_i = (S_1^i, S_2^i, S_3^i, S_4^i)$ is a SFN

1) Calculate the center of SFN $P_i$ as

$$C_i = \frac{S_2^i + S_3^i}{2} \tag{29}$$

2) Get the non-specificity for SFN $P_i$ as the following

$$H_i = log_2(S_3^i - S_2^i + 1) \tag{30}$$

3) Get the fuzziness for SFN

The fuzziness of left and right shadow intervals $f_l$ and $f_r$ for every SFN as follows.

$$f_l = S_2^i - S_1^i, \tag{31}$$

$$f_r = S_4^i - S_3^i \tag{32}$$

and the fuzziness $f_i$ of SFN $P_i$ calculates as

$$f_i = f_l + f_r \tag{33}$$

4) Calculate the rank index $R_i$ of SFN $P_i$

$$R_i = \frac{C_i}{H_i + f_i} \tag{34}$$

## 4. NUMERICAL EXAMPLE

In this section, we will apply a new approach to select the best supplier of ready-mixed concrete for a project located in a large city. The following example from [32] is shown in Figure 6. Three suppliers' alternatives $(A_1, A_2, A_3)$ produce the mix in accordance with a precise specification. Four criteria are used to evaluate the qualification of the supplier: the price of the mix $(C_1)$, the location of the batching plant $(C_2)$, the terms of payment $(C_3)$, and the batching plant's capacity $(C_4)$. The pairwise comparisons of the criteria's relative importance matrix values vary between exact values, interval numbers (INs), type-l fuzzy numbers and intuitionistic fuzzy numbers (IFNs), as in Tables 2 and 3. Also, two different types of fuzzy numbers are used to compare between criteria, i.e. triangular (TFNs) and Gaussian fuzzy numbers (GFNs). We apply a new method to rank the suppliers using the following steps:

Step 1: Pairwise comparison matrices between alternatives for each criterion are created as in Tables 4, 5, 6, and 7. These matrices' values are represented in the form of crisp numbers, TFNs, GFNs and IFNs. The consistency ratio is calculated by using crisp numbers in Table 1, and the result is less than the threshold value of 0.1.

Step 2: Transform multi-granular values of matrices to shadowed fuzzy numbers (SFNs), as in Tables 8, 9, 10, 11, and 12. We use the (6), (7), (8), (9), and (10) and (13), (14), (15), (16), (17), (18), (19), and (20).

Step 3: The SFN geometric mean value for each criterion is calculated according to (26) from Table 8. Also, the same mean is computed for each pairwise comparison matrix between alternatives from Tables 9, 10, 11, and 12.

Step 4: The normalized weight of each criterion is calculated according to (27). The results are shown in Table 13.

Step 5: The normalized priority for every alternative according to the particular criterion is determined using the (26) and (27) and the results are illustrated in Table 14.

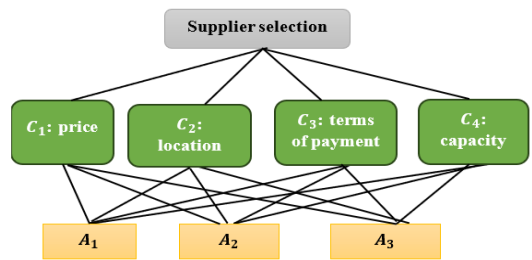

Figure 6. The problem hierarchical

Step 6: The final priority or preference value of each alternative is calculated using (28) and the result is shown in Table 15.

Step 7: To rank the final preference SFNs, we obtained the rank index using (29),(30), (31), (32), (33), and (34) as the following : $R(A_1) = 0.221$, $R(A_2) = 0.372$ and $R(A_3) = 0.4242$. These values of ranking index indicate that the order of suppliers is $A_3 > A_2 > A_1$.

Table 2. The pairwise comparison between criteria by using muti-granular numbers

|       | $C_1$    | $C_2$     | $C_3$    | $C_4$     |
|-------|----------|-----------|----------|-----------|
| $C_1$ | 1        | TFN(1/5)  | IN(3)    | GFN(1/3)  |
| $C_2$ | TFN(5)   | 1         | IFN(5)   | TFN(1)    |
| $C_3$ | IN(1/3)  | IFN(1/5)  | 1        | GFN(1/3)  |
| $C_4$ | GFN(3)   | TFN(1/1)  | GFN(3)   | 1         |

Table 3. The pairwise comparison between criteria according to Tables 1 and 2

|       | $C_1$      | $C_2$                        | $C_3$              | $C_4$       |
|-------|------------|------------------------------|--------------------|-------------|
| $C_1$ | 1          | (1/7, 1/5, 1/3)              | [1, 5]             | (1/3, 0.9)  |
| $C_2$ | (3, 5, 7)  | 1                            | (4,5,6),(3,5,7)    | (1, 1, 2)   |
| $C_3$ | [1/5, 1]   | (1/6, 1/5, ¼), (1/7,1/5,1/3) | 1                  | (1/3, 0.9)  |
| $C_4$ | (3, 0.9)   | (1/2,1,1)                    | (3, 0.9)           | 1           |

Table 4. The pairwise comparison between alternatives for the price criterion

|  | $A_1$ | $A_2$ | $A_3$ |
|---|---|---|---|
| $A_1$ | 1 | (1/7, 0.9) | (1/4,1/3,1/2),(1/5,1/3, 1) |
| $A_2$ | (7, 0.9) | 1 | (5, 7, 9) |
| $A_3$ | (2,3,4) ,(1,3,5) | (1/9,1/7,1/5) | 1 |

Table 5. The pairwise comparison between alternatives for the location criterion

|  | $A_1$ | $A_2$ | $A_3$ |
|---|---|---|---|
| $A_1$ | 1 | (1/3, 09) | (1/9.5,1/9,1/8), (1/10,1/9,1/7) |
| $A_2$ | (3, 09) | 1 | (1/7,1/5,1/3) |
| $A_3$ | ( 8,9,9.5), (7,9,10) | (3,5,7) | 1 |

Table 6. The pairwise comparison between alternatives for the payment criterion

|  | $A_1$ | $A_2$ | $A_3$ |
|---|---|---|---|
| $A_1$ | 1 | (1/7, 0.9) | (6,7,8), (5,7,9) |
| $A_2$ | (7, 0.9) | 1 | (8,9,10) |
| $A_3$ | (1/8,1/7,1/6) ,(1/9,1/7,1/5) | (1/10,1/9,1/8) | 1 |

Table 7. The pairwise comparison between alternatives for the capacity criterion

|  | $A_1$ | $A_2$ | $A_3$ |
|---|---|---|---|
| $A_1$ | 1 | (1/5, 0.9) | (1/6,1/5,1/4) , (1/7,1/5,1/3) |
| $A_2$ | (5, 0.9) | 1 | (1,1,2) |
| $A_3$ | (4,5,6), (3,5,7) | (1/2,1,1) | 1 |

Table 8. The pairwise comparison between criteria using SFNs

|  | $C_1$ | $C_2$ | $C_3$ | $C_4$ |
|---|---|---|---|---|
| $C_1$ | (1, 1, 1, 1) | (0.14, 0.17, 0.27, 0.33) | [1, 5] | (0, 0, 1.06, 1.996) |
| $C_2$ | (3.12, 4.12, 5.88, 6.88) | (1, 1, 1, 1) | (3.655, 4.35, 5.65, 6.35) | (0.97, 0.97, 0.97, 1) |
| $C_3$ | [0.2, 1] | (0.16, 0.18, 0.25, 0.3) | (1, 1, 1, 1) | (0, 0, 1.06, 1.996) |
| $C_4$ | (1.33, 2.27, 3.73, 4.67) | (0.51, 0.76, 1, 1) | (1.33, 2.27, 3.73, 4.64) | (1, 1, 1, 1) |

Table 9. The pairwise comparison between alternatives for the price criterion using SFNs

|  | $A_1$ | $A_2$ | $A_3$ |
|---|---|---|---|
| $A_1$ | (1, 1, 1, 1) | (0, 0, 0.87, 0.81) | (0.24, 0.29, 0.54, 1.01) |
| $A_2$ | (5.33, 6.27, 7.73, 8.67) | (1, 1, 1, 1) | (5.12, 6.12, 7.88, 8.88) |
| $A_3$ | (1.66, 2.35, 3.65 , 4.35) | (0.11, 0.13, 0.17, 0.2) | (1, 1, 1, 1) |

Table 10. The pairwise comparison between alternatives for the location criterion using SFNs

|  | $A_1$ | $A_2$ | $A_3$ |
|---|---|---|---|
| $A_1$ | (1, 1, 1, 1) | (0, 0, 1.06, 1.996) | (0.106, 0.11, 0.12, 0.13) |
| $A_2$ | (1.33, 2.27, 3.73, 4.67) | (1, 1, 1, 1) | (0.14, 0.17, 0.27, 0.33) |
| $A_3$ | (7.65, 8.34, 9.33, 9.68) | (3.12, 4.12, 5.88, 6.88) | (1, 1, 1, 1) |

Table 11. The pairwise comparison between alternatives for the payment criterion using SFNs

|  | $A_1$ | $A_2$ | $A_3$ |
|---|---|---|---|
| $A_1$ | (1, 1, 1, 1) | (0, 0, 0.87, 1.81) | (5.655, 6.35, 7.65, 8.35) |
| $A_2$ | (5.33, 6.27, 7.73, 8.67) | (1, 1, 1, 1) | (8.05, 8.55, 9.45, 9.95) |
| $A_3$ | (0.1187, 0.13, 0.16, 0.18) | (0.09, 0.1, 0.12, 0.13) | (1, 1, 1, 1) |

Table 12. The pairwise comparison between alternatives for the capacity criterion using SFNs

|  | $A_1$ | $A_2$ | $A_3$ |
|---|---|---|---|
| $A_1$ | (1, 1, 1,1) | (0, 0, 0.93, 1.87) | (0.16, 0.18, 0.25, 0.304) |
| $A_2$ | (3.33, 4.27, 5.73, 6.67) | (1, 1, 1, 1) | (0.97, 0.97, 0.97, 1) |
| $A_3$ | (3.655, 4.35, 5.65, 6.35) | (0.51, 0.76, 1, 1) | (1, 1, 1, 1) |

Table 13. The normalized weights of criteria

| $C_1$ | $C_2$ | $C_3$ | $C_4$ |
|---|---|---|---|
| (0, 0, 0.3171, 0.322) | (0.28, 0.33, 0.69, 0.92) | (0, 0, 0.21, 0.32) | (0.15, 0.23, 0.56, 0.77) |

Table 14. The normalized priority of for every alternative according to the particular criterion

|  | $A_1$ | $A_2$ | $A_3$ |
|---|---|---|---|
| $C_1$ | ( 0, 0, 0.19, 0.34) | (0.47, 0.61, 0.97, 1.19) | (0.09, 0.12, 0.21, 0.27) |
| $C_2$ | (0, 0, 0.13, 0.19) | (0.098, 0.14, 0.25, 0.34) | (0.49, 0.61, 0.96, 1.18) |
| $C_3$ | (0, 0, 0.47, 0.66) | (0.49, 0.596, 1.04, 1.19) | (0.03, 0.04, 0.07, 0.08) |
| $C_4$ | (0, 0, 0.199, 0.31) | (0.32, 0.39, 0.57, 0.695) | (0.27, 0.36, 0.58, 0.68) |

Table 15. The final preference values of alternatives

| $A_1$ | $A_2$ | $A_3$ |
|---|---|---|
| (0, 0, 0.36, 0.73) | (0.076, 0.13, 1.02, 1.6) | (0.18, 0.29, 1.06, 1.72) |

## 5. DISCUSSION

According to the previous steps, and comparing them with Alsawy et al. [16], Chatterjee et al. [17] , and Aydin et al. [18] the following remarks are found:
1) With the methods Alsawy et al. [16], and Chatterjee et al. [17], G-numbers represent a higher level of abstraction but do not consider different types of uncertainty which the current method outperforms.

2) In the Aydin et al. approach [18], transform only crisp, interval, and type-1 fuzzy numbers. The newly proposed approach converts the previous types in addition to the higher types of fuzzy numbers.

3) The new approach uses the uncertainty properties of aggregation preference values of alternatives to rank them.

4) Solving the same illustrative example adopted by Biruk et al. [32], using the proposed approach results in the same ranking of alternatives even though different types of granular information are used. This demonstrates the validity of the proposed method.

## 6. CONCLUSIONS

In multi-granular AHP problems, different types of uncertain numbers are used to express relative preferences between criteria and similarly with alternatives. To solve this pattern of AHP problems, a new algorithm called shadowed AHP is proposed. It converts different types of numbers, including crisp, interval, type-1 fuzzy, and intuitionistic fuzzy numbers, to shadowed fuzzy numbers. The new shadowed AHP steps have been developed. Also, a new ranking method is proposed to rank the results of the evaluation values of alternatives, which are represented in the form of shadowed fuzzy numbers. The new algorithm is applied to multi-attribute supplier selection decision-making problems. The new technique unifies the form of linguistic terms and, at the same time, preserves the features of uncertainty in input information. It is superior to previous methods that transformed different types of granular numbers into unified uncertain numbers with preserved uncertainty patterns in the input multi-granular information. Also, it can be used with higher types of fuzzy numbers. The rank of alternatives is based on their uncertainty features.